\documentclass[11pt]{article}

\usepackage[final]{acl}

\usepackage{times}
\usepackage{latexsym}
\usepackage{booktabs}
\usepackage{multirow}
\usepackage{amsmath}
\usepackage{amssymb}
\usepackage{paralist}
\usepackage{stmaryrd}
\usepackage{paralist}
\usepackage[ruled,vlined]{algorithm2e}
\usepackage{todonotes}
\usepackage{xurl}
\usepackage{hyperref} 
\usepackage[table]{xcolor} 
\usepackage{makecell}

\usepackage[T1]{fontenc}
\usepackage{tcolorbox}

\usepackage[utf8]{inputenc}

\usepackage{microtype}

\usepackage{inconsolata}
\usepackage{textcomp}
\usepackage{pifont}
\usepackage{graphicx}

\title{Neither Over- nor Under-Constrained: \\ Evaluating LLM Personalization under a Constraint Qualification Audit}

\author{
 \textbf{Xuran Li\textsuperscript{1}},
 \textbf{Guanqin Zhang\textsuperscript{1}},
 \textbf{Imran Razzak\textsuperscript{1,2}},
 \\
 \textbf{Hakim Hacid\textsuperscript{3}},
 \textbf{Eleanna Kafeza\textsuperscript{3}},
 \textbf{Hao Xue\textsuperscript{1,4}},
 \textbf{Flora D. Salim\textsuperscript{1}},
\\
 \textsuperscript{1}University of New South Wales,
 \\
 \textsuperscript{2}Mohamed bin Zayed University of Artificial Intelligence,
\\
 \textsuperscript{3}The Technology Innovation Institute,
 \\
 \textsuperscript{4}The Hong Kong University of Science and Technology
}

\begin{document}
\maketitle
\begin{abstract}
Accurate evaluation of LLM personalization is hindered by current evaluation paradigms that over- or under-constrain the response space, yielding frequent false rejections and false acceptances. To reduce these false rates, we introduce ConQual-Audit, an audit-based framework that evaluates LLM personalization through a three-stage process: (1) extracting semantic constraints from user profiles and task-grounded reference; (2) qualifying constraints by pruning unnecessary constraints and recovering missing constraints; and (3) verifying candidate responses against the audited constraints via natural language inference. Across PrefEval and LaMP, ConQual-Audit outperforms nine baselines, attaining top AUCs (0.88 and 0.90) and the lowest false rate (0.21) on the leanest constraint sets at an amortizable audit cost. Crucially, this constraint audit is portable: integrating it into ExPerT and PREF helps revise smaller, higher-quality constraint sets to reduce their evaluation false rates. All code, prompts, and evaluation scripts are released in an anonymous repository \cite{code_anonymous_repo}

\end{abstract}

\section{Introduction}
LLMs demonstrate strong general-purpose capabilities across a wide range of tasks \cite{minaee2025largelanguagemodelssurvey, niu2025largelanguagemodelscognitive}. Beyond generic performance, personalization has emerged as a crucial requirement for enhancing user satisfaction and enabling context-aware interactions. Specifically, personalization refers to tailoring LLM outputs to diverse user preferences, contextual signals, and implicit intents \cite{PersonalizationSurvey}. 

\begin{figure}[t]
    \centering
    \includegraphics[width=1.\linewidth]{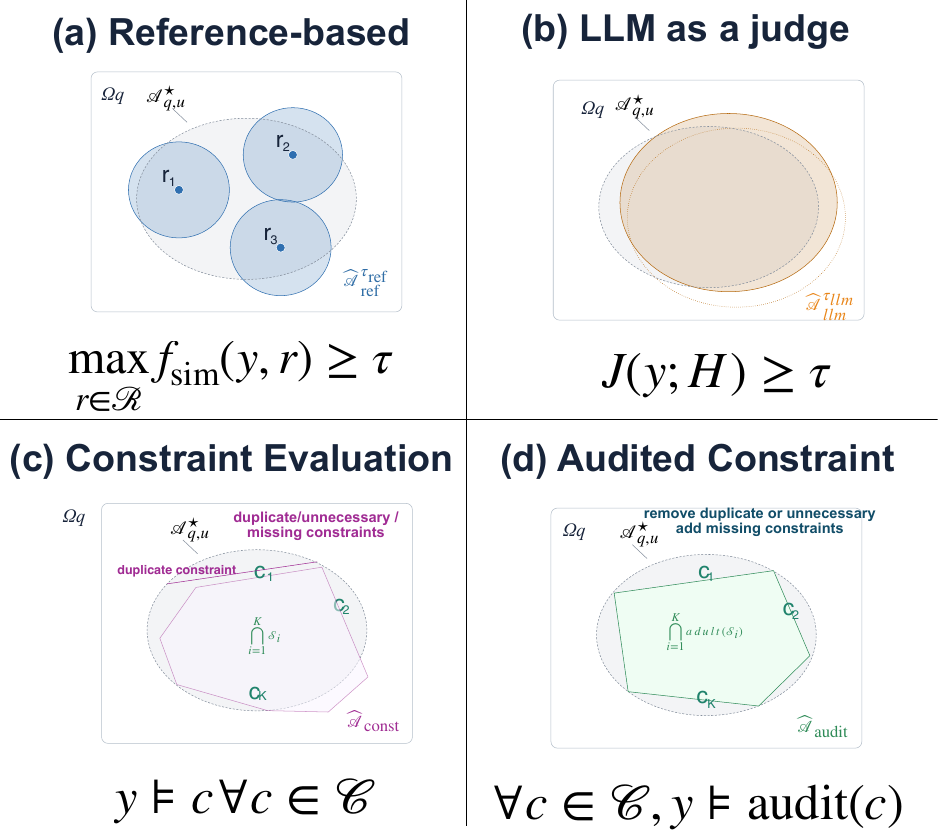}
    \caption{Personalized Response Space. }
    \label{pic:PersonalizedResponseSpace}
\end{figure}

However, current evaluation paradigms fail to reliably evaluate LLM personalization, as they tend to either over-constrain or under-constrain the personalized response space, prematurely penalizing valid outputs or tolerating invalid ones (Figure~\ref{pic:PersonalizedResponseSpace}). On one hand, reference-based metrics (e.g., BLEU~\cite{papineni-etal-2002-bleu}, ROUGE ~\cite{dasgupta2024persevalassessingpersonalizationtext}, BERTScore~\cite{bertscore}, MoverScore~\cite{MoverScore}) depend on proximity to rigid human references; they reject acceptable semantic variations while erroneously tolerating outputs with high surface overlap but critical personalization errors. On the other hand, LLM-as-a-judge frameworks (e.g., FactScore~\cite{factscore}, GEMBA~\cite{kocmi-federmann-2023-large}, PerSE~\cite{wang-etal-2024-learning-personalized}) can reason over flexible outputs, but they function as black-box evaluators that lack transparent and interpretable evaluation mechanisms. While recent constraint protocols (e.g., PREF~\cite{fu2025prefreferencefreeevaluationpersonalised}, ExPerT~\cite{salemi2025expert}) enhance interpretability via a two-stage paradigm—extracting constraints from user profiles and assessing response adherence—they remain brittle: synthesized constraints often suffer from low fidelity, manifesting as redundancy, omissions, or hallucinated requirements.

\begin{figure*}[h!]
    \centering
    \includegraphics[width=1.\linewidth]{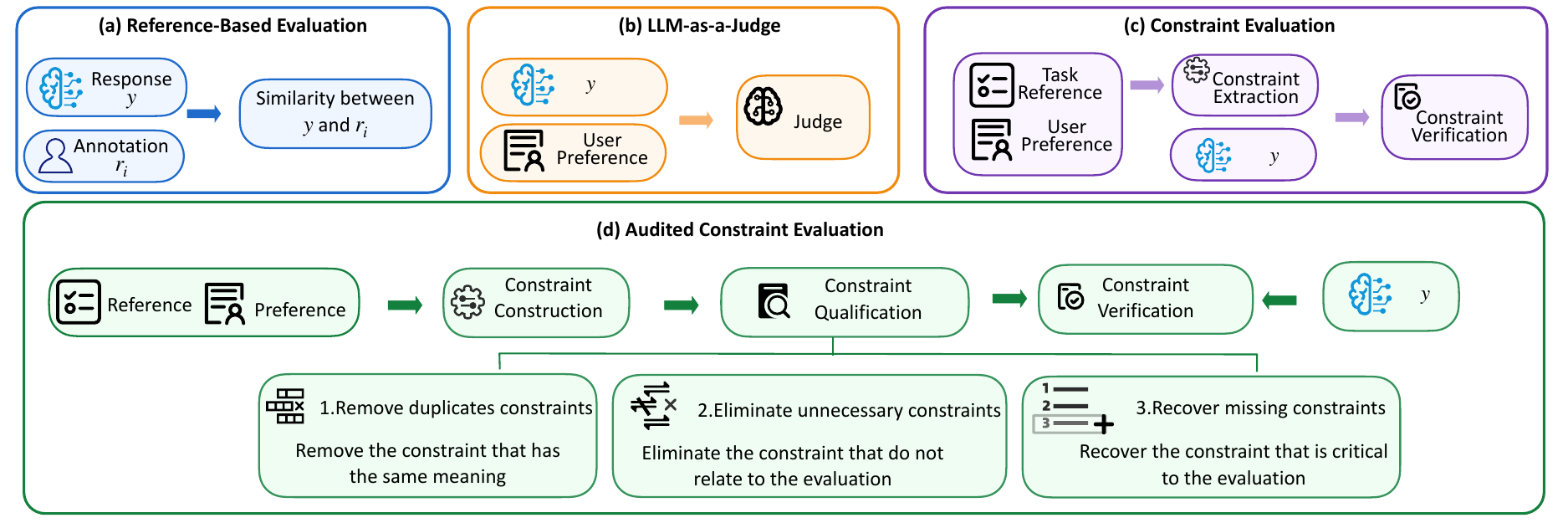}
    \caption{Evalauation pipeline of ConQual-Audit framework }
    \label{fig:overview}
\end{figure*}

To address both over- and under-constraining problems, we introduce ConQual-Audit, an audit-based framework that establishes calibrated, interpretable boundaries over the personalized response space via a principled constraint qualification mechanism (Figure~\ref{fig:overview}). {ConQual-Audit} operates across three modular stages: \emph{First}, it extracts original semantic constraints from both task-grounded reference and user preference. \emph{Second}, it refines the extracted constraints through a counterexample-driven tripartite audit: (i) deduplicating redundant criteria; (ii) pruning unnecessary constraints by constructing personalized counterexamples that violate candidate criteria yet remain valid; and (iii) recovering missing constraints by synthesizing non-personalized counterexamples that satisfy all active criteria. \emph{Finally}, it evaluates candidate responses against the audited constraint set via a premise–hypothesis inference formulation, and unifies constraint satisfaction and response attribution via their harmonic mean to jointly balance constraint coverage and response recall.

\emph{Empirical Verification.} 
Extensive empirical evaluations across the {PrefEval} \citet{zhao2025llmsrecognizepreferencesevaluating} and {LaMP} \cite{LaMP} benchmarks show the effectiveness of {ConQual-Audit}. Our framework establishes the highest evaluation performance, achieving top AUC scores ($0.88$ and $0.90$) and the lowest overall false rates ($\text{FR} = 0.21$) while operating on the most compact constraint sets ($2.36$ and $2.97$ per query). Crucially, we show that constraint qualification is model-agnostic and highly portable: integrating our audit module into existing evaluators ({ExPerT} and {PREF}) systematically remove unnecessary constraints, restore missing ones, and significantly reduce downstream evaluation errors at an amortized audit computational cost.

The primary contributions are three-fold:
\begin{itemize}
    \item \textbf{{ConQual-Audit} Framework:} We propose a principled, three-stage audit-based framework for LLM personalization evaluation that overcomes both over- and under-constrained evaluation regimes through semantic constraint construction, qualification, and verification.
    \item \textbf{Counterexample-Guided Qualification:} We design a tripartite audit protocol that systematically deduplicates, prunes, and recovers constraints. This mechanism yields compact, calibrated constraint sets that attain top AUCs and the lowest overall false rates on {LaMP} and {PrefEval}.
    \item \textbf{Modularity and Portability:} We demonstrate that constraint qualification is model-agnostic: augmenting external constraint-based evaluators ({ExPerT} and {PREF}) with our audit module consistently improves constraint quality and lowers downstream false rates with minimal overhead.
\end{itemize}

\section{Problem Formulation \& Related Work}
\label{sec:problem_formulation}

\paragraph{Problem formulation.}
We formalize LLM personalization evaluation \cite{LaMP-QA,LongLaMP,PersonalizationSurvey,LLMPersonalizedJudge} as a set-membership problem. Let $q$ be a query from a user with interaction history $H$, let $y^\ast$ be a task reference for $q$, and let $\Omega$ be the set of all candidate responses to $q$. The user and the task determine the set of acceptable responses
\begin{equation}
  \mathcal{A}^{\star} = \{\, y \in \Omega : y \text{ is acceptable} \,\},
  \label{eq:latent_acceptable_set}
\end{equation}
where a response is acceptable if it follows the user's preferences, as evidenced by $H$, and meets the task requirements specified by $y^\ast$. Evaluating a response $y$ means deciding whether $y \in \mathcal{A}^{\star}$. Since $\mathcal{A}^{\star}$ is not observed, every evaluation method approximates it by an induced set $\widehat{\mathcal{A}}$; existing methods differ in how they construct it (Figures~\ref{pic:PersonalizedResponseSpace} and~\ref{fig:overview}).

\paragraph{Reference-based evaluation}
(e.g., BLEU~\cite{papineni-etal-2002-bleu}, ROUGE~\cite{dasgupta2024persevalassessingpersonalizationtext}, BERTScore~\cite{bertscore}, MoverScore~\cite{MoverScore}) accepts responses that are similar to a set of human annotations $\mathcal{R} \subset \mathcal{A}^{\star}$ via a similarity function $\mathrm{sim}$ and a threshold $\tau$:
\begin{equation}
  \widehat{\mathcal{A}}_{\mathrm{ref}} = \{\, y \in \Omega : \max_{r \in \mathcal{R}} \mathrm{sim}(y, r) \geq \tau \,\},
  \label{eq:reference_induced_set}
\end{equation}
Although annotations serve as effective anchors, they cannot encompass all forms of personalized responses. Consequently, they often reject valid responses that differ in wording from the annotations, while accepting invalid responses that happen to share similar vocabulary (such as negated answers).


\paragraph{LLM-as-a-judge}
(e.g., GEMBA~\cite{kocmi-federmann-2023-large}, FActScore~\cite{factscore}, PerSE~\cite{wang-etal-2024-learning-personalized}) asks an LLM $J$ \cite{llama3modelcard, deepseekai2025deepseekr1incentivizingreasoningcapability, qwen3technicalreport,falcon-h1r} to score a response given the user's history:
\begin{equation}
  \widehat{\mathcal{A}}_{\mathrm{llm}} = \{\, y \in \Omega : J(y; H) \geq \tau \,\}.
  \label{eq:judge_induced_set}
\end{equation}
LLM judges are not tied to references, but they score a response holistically, so the boundary of $\widehat{\mathcal{A}}_{\mathrm{llm}}$ is implicit in the judge and can be neither inspected nor audited.

\paragraph{Constraint evaluation}
(e.g., ExPerT~\cite{salemi2025expert}, PREF~\cite{fu2025prefreferencefreeevaluationpersonalised}) describes $\mathcal{A}^{\star}$ by explicit constraints, such as rubric criteria in PREF or content aspects in ExPerT. They check responses against a constraint set $\mathcal{C}$ and induce:
\begin{equation}
  \widehat{\mathcal{A}}(\mathcal{C}) = \{\, y \in \Omega : y \models c \forall c \in \mathcal{C} \,\},
  \label{eq:constraint_defined_response_set}
\end{equation}
where $y \models c$ denotes that $y$ satisfies the constraint $c$; in practice, the conjunction is relaxed to a graded score. Since the constraints specify the required content of a response rather than the exact wording, this approach accepts a variety of valid responses, and evaluation results can be traced back to the underlying constraints.

However, the reliability of constraint-based evaluation depends on the quality of the constraints themselves; existing constraint-based evaluators often lack mechanisms to review these constraints during use, which can lead to the false acceptance of invalid responses or the false rejection of valid ones due to unqualified constraints.

\section{Constraint Quality Analysis}
\label{sec:constraint-impact}

A constraint evaluator errs when $\widehat{\mathcal{A}}(\mathcal{C})$ differs from $\mathcal{A}^{\star}$ (Figure~\ref{pic:PersonalizedResponseSpace}). The difference consists of two error kinds: 1). False rejection $\mathrm{FR}(\mathcal{C})$ (i.e., the rejection of valid responses, constituting over-constraint); 2). False Acceptance $\mathrm{FA}(\mathcal{C})$ (i.e., the acceptance of invalid responses, constituting under-constraint). 
\begin{equation}
  \mathrm{FR}(\mathcal{C}) = \mathcal{A}^{\star} \setminus \widehat{\mathcal{A}}(\mathcal{C}),
  \mathrm{FA}(\mathcal{C}) = \widehat{\mathcal{A}}(\mathcal{C}) \setminus \mathcal{A}^{\star}.
  \label{eq:error-sets}
\end{equation}

\subsection{Necessity and Sufficiency}
\label{subsec:boundary-criteria}

$\widehat{\mathcal{A}}(\mathcal{C})$ is the intersection of all constraints, to reduce over-constraint and under-constraint problems, constraints should possess these two properties:

\emph{1. Individual necessity (no false rejections).}  $\mathrm{FR}(\mathcal{C}) = \emptyset$ holds if and only if every constraint is satisfied by every valid response:
\begin{equation}
  y \models c, \forall y \in \mathcal{A}^{\star} \text{ and } \forall c \in \mathcal{C}.
  \label{eq:necessity}
\end{equation}
A constraint that some valid response violates is \emph{unnecessary}; each unnecessary constraint adds its own false rejections.

\emph{2. Joint sufficiency (no false acceptances).} $\mathrm{FA}(\mathcal{C}) = \emptyset$ holds if and only if the constraints together admit only acceptable responses:
\begin{equation}
  \widehat{\mathcal{A}}(\mathcal{C}) \subseteq \mathcal{A}^{\star}.
  \label{eq:sufficiency}
\end{equation}
Unlike necessity, this property is for the whole set. A set that violates it is \emph{insufficient}, because a necessary constraint that would exclude a falsely accepted response is \emph{missing}.

Necessity gives $\mathcal{A}^{\star} \subseteq \widehat{\mathcal{A}}(\mathcal{C})$ and sufficiency the reverse inclusion, they together recover the acceptable set exactly.

\subsection{Constraint Quality Impact on Evaluation}
\label{sec:cq-score}

The constraint evaluators compute the satisfaction rate $s_{\mathcal{C}}(y)$ and accept a response if $s_{\mathcal{C}}(y) \ge \tau$:
\begin{equation}
  s_{\mathcal{C}}(y) = \frac{|\{c \in \mathcal{C} : y \models c\}|}{|\mathcal{C}|}.
  \label{eq:score}
\end{equation}
Weighted criteria, as in PREF, follow an analogous continuous formulation. Let $n = |\mathcal{C}|$ and $s = s_{\mathcal{C}}(y)$. Modifying $\mathcal{C}$ by adding a constraint ($+c$) or removing one ($-c$) perturbs the score by:
\begin{equation}
  \Delta s_{-} = \frac{s - \mathbb{I}(y \models c)}{n - 1}
  \Delta s_{+} = \frac{\mathbb{I}(y \models c) - s}{n + 1}
  \label{eq:score-shift}
\end{equation}
The direction and magnitude of this perturbation depend strictly on whether $y$ satisfies $c$ relative to its baseline score $s$. For a valid response $y^{+} \in \mathcal{A}^{\star}$ and an invalid response $y^{-} \notin \mathcal{A}^{\star}$, this establishes three theoretical links between constraint quality and evaluation error:

\emph{1. Unnecessary constraints inflate false rejections ($\mathrm{FR}(\mathcal{C}) \uparrow$).}
When a constraint $c$ violates individual necessity (Eq.~\eqref{eq:necessity}), valid responses may violate it ($\mathbb{I}(y^{+} \models c) = 0$). Adding $c$ penalizes the valid response by $\Delta s_{+} = -s/(n + 1) \le 0$. If $s - \Delta s_{+} < \tau$, this induces a false rejection. Conversely, pruning unnecessary constraints via qualification guarantees $\Delta s_{-} = s/(n - 1) \ge 0$, directly restoring valid responses above $\tau$.

\emph{2. Insufficient sets inflate false acceptances ($\mathrm{FA}(\mathcal{C}) \uparrow$).}
If $\mathcal{C}$ violates joint sufficiency (Eq.~\eqref{eq:sufficiency}), a necessary constraint $c^{\star}$ that $y^{-}$ violates ($\mathbb{I}(y^{-} \models c^{\star}) = 0$) is omitted. The absence of $c^{\star}$ inflates the invalid response's score by $\Delta s_{-} = s/(n - 1) \ge 0$. Omitting necessary constraint can decrease the threshold distance for invalid responses, leading to false acceptances.


\begin{algorithm}[h]
\caption{ConQual-Audit}
\label{alg:conqual_p}
\DontPrintSemicolon
\LinesNumbered
\SetCommentSty{algcommentfont}
\KwIn{ $q$,  $H$,  $y^\ast$,  $\mathcal{Y}$;  $G$,  $J$, $V$}
\KwOut{Scores $\{(R_y, P_y, F1_y)\}_{y \in \mathcal{Y}}$}
\tcp{1. Construction}
$\mathcal{C}^{p} \gets G(q, H)$;\ \ $\mathcal{C}^{a} \gets G(q, y^\ast)$\;
\tcp{2. Qualification (Alg.~\ref{alg:qualify})}
$\mathcal{C} \gets \textsc{Qualify}(\mathcal{C}^{p} \cup \mathcal{C}^{a})$\;
\tcp{3. Verification}
\ForEach{$y \in \mathcal{Y}$}{
  $\mathcal{C}_y^{E} \gets \{c \in \mathcal{C} : V(y, c) = \textsc{E}\}$\;
  $\{y_c^{E}\}_{c \in \mathcal{C}_y^{E}} \gets V(S(y), \mathcal{C}_y^{E})$\;
  $R_y \gets |\mathcal{C}_y^{E}| \,/\, |\mathcal{C}|$\;
  $P_y \gets \mathrm{tok}\big(\bigcup_{c} y_c^{E}\big) \,/\, \mathrm{tok}\big(S(y)\big)$\;
  $F1_y \gets 2 P_y R_y \,/\, (P_y + R_y)$\;
}
\Return{$\{(R_y, P_y, F1_y)\}_{y \in \mathcal{Y}}$}
\end{algorithm}

\section{The ConQual-Audit Framework}
\label{sec:framework}
A low-quality constraint set can lead to increased false rejections and false acceptances. Therefore, it is crucial to audit constraints before constraint validation. ConQual-Audit is proposed to evaluate LLM personalization under audited constraints. 
As shown in ALG.~\ref{alg:conqual_p}, ConQual-Audit proceeds in three stages:
(1)~\emph{construction} (\S\ref{subsec:construction}) derives original constraints from user preference and task reference;
(2)~\emph{qualification} (\S\ref{subsec:qualification}) audits and improves necessity and sufficiency of the original constraints (\S\ref{subsec:boundary-criteria}); 
and (3)~\emph{verification} (\S\ref{subsec:verification}) checks response against the audited constraints and computes evaluation result scores. 

\subsection{Constraint Construction}
\label{subsec:construction}

Each constraint is designed as a declarative proposition about the response, and $y \models c$ holds when $y$ entails $c$. A required $c$ is stated as a positive assertion (e.g., \emph{``The response recommends whole-food sources of protein.''}), which is entailed only if the response provides the required content. A prohibited $c$ is stated as a negative assertion (e.g., \emph{``The response does not recommend protein powder.''}), which is entailed when the response omits or explicitly rejects the corresponding content.  Therefore, the constraints can be verified by natural language inference, during the constraint verification stage.

As shown in ALG.~\ref{alg:conqual_p}, preference constraints and answer constraints are constructed by a generator $G$ (line 1). Specifically, $G$ extracts the user's interests from $H$, and derives one constraint for each thing the user prefers or avoids among the interests that apply to the query $q$. For the answer constraint, $G$ decomposes task-grounded reference $y^\ast$ into atomic requirements, covering what an valid answer must provide and what it must avoid.

\subsection{Constraint Qualification}
\label{subsec:qualification}
Constraint qualification audits individual necessity (Eq.~\ref{eq:necessity}) and joint sufficiency (Eq.~\ref{eq:sufficiency}). Since $\mathcal{A}^{\star}$ is not observed, it tests them with counterexamples (ALG.~\ref{alg:qualify}): the generator $G$ produces a response $\tilde{y}$ subject to preset constraints; the verifier $V$ confirms that the constraints are satisfied; and the discriminator $J$ determines whether $\tilde{y}$ belongs to $\mathcal{A}^{\star}$ by verifying it against user preferences or task reference expectations. If no valid counterexample is found in $K$ attempts, the constraints is unchanged.

\begin{algorithm}[h]
\caption{Constraint Qualification.}
\label{alg:qualify}
\DontPrintSemicolon
\LinesNumbered
\SetCommentSty{algcommentfont}
\SetFuncSty{textsc}
\SetArgSty{textup}
\SetKwFunction{FW}{Adverse}
\SetKwFunction{ADD}{Update}
\KwIn{Constraints $\mathcal{C}$; $q, H, y^\ast$ and $G, J, V$}
\KwOut{Qualified constraints $\mathcal{C}$}
\tcp{Duplicated}
\lForEach{$i < j$ \KwSty{with} $c_i \equiv c_j$}{$\mathcal{C} \gets \mathcal{C} \setminus \{c_j\}$}
\tcp{Unnecessary (necessity)}
\ForEach{$c \in \mathcal{C}$, \KwSty{while} $|\mathcal{C}| > 1$}{
  $\tilde{y} \gets \FW(\mathcal{C} \setminus \{c\}, \{c\}, \cdot)$\;
  \lIf{$\tilde{y} \neq \bot \wedge J(\tilde{y}) = \textsc{Pass}$}{$\mathcal{C} \gets \mathcal{C} \setminus \{c\}$}
}
\tcp{Missing (sufficiency)}
\ForEach{$r \in \{\textsc{Ref}, \textsc{Pref}\}$}{
  $\tilde{y} \gets \FW(\mathcal{C}, \emptyset, r)$\;
  \If{$\tilde{y} \neq \bot \wedge J(\tilde{y}) = \textsc{Fail}$}{
    $\mathcal{C} \gets \mathcal{C} \cup \ADD(\mathcal{C}, \tilde{y})$\;
  }
}
\Return{$\mathcal{C}$}\;
\end{algorithm}


Specifically, Alg.~\ref{alg:qualify} first applies a duplicated constraints check in line 1. If two constraints are duplicates, $c_i \equiv c_j$, the later one is removed. This does not change $\widehat{\mathcal{A}}(\mathcal{C})$ but simplifies the constraints for following audits. Lines~2--4 audit unnecessary constraints (lines~2--4). For each $c$, $G$ generates a $\tilde{y}$ that satisfies every other constraint but violates $c$. If $J$ accepts it, then $\tilde{y} \in \mathcal{A}^{\star}$ and $\tilde{y} \not\models c$, so $c$ violates necessity and is removed. Each constraint is tested against those still kept. Finally, lines~5--8 imply missing constraints audit. For the task reference and user preference in turn, $G$ generates a $\tilde{y}$ that satisfies every constraint but breaks that exceptions. If $J$ rejects it, then $\tilde{y} \in \mathrm{FA}(\mathcal{C})$, which violates sufficiency. $G$ then proposes new constraints from the broken requirements and $\tilde{y}$, and a proposal is added to exclude $\tilde{y}$ from other valid responses.

\subsection{Constraint Verification and Metrics}
\label{subsec:verification}

Given the qualified set $\mathcal{C}$, $V$ checks a response in two steps in ALG.~\ref{alg:conqual_p}. (1)~\emph{Response verification (line 4).} With $y$ as the premise and each constraint as a hypothesis, $V$ labels every constraint as \textsc{Entailment}, \textsc{Contradiction} or \textsc{Neutral}. The satisfied constraints are $\mathcal{C}_y^{E} = \{c \in \mathcal{C} : y \models c\}$. (2)~\emph{Evidence attribution (line 5).} For each $c \in \mathcal{C}_y^{E}$, $V$ identifies the sentences $y_c^{E} \subseteq S(y)$ that support it, where $S(y)$ are the sentences of $y$. Then, lines~6--8 compute the evaluation results: 
\begin{equation}
\begin{gathered}
  R_y = \frac{|\mathcal{C}_y^{E}|}{|\mathcal{C}|}, \quad
  P_y = \frac{\mathrm{tok}\big(\bigcup_{c \in \mathcal{C}_y^{E}} y_c^{E}\big)}{\mathrm{tok}\big(S(y)\big)},\\
  F1_y = \frac{2 P_y R_y}{P_y + R_y},
\end{gathered}
  \label{eq:metrics}
\end{equation}
where $\mathrm{tok}(\cdot)$ counts word tokens. 

The constraint recall $R_y$ is the score $s_{\mathcal{C}}(y)$ of Eq.~\eqref{eq:score}: it measures how many constraints the response meets, and a response is accepted if $R_y \geq \tau$. The response precision $P_y$ measures the share of the response devoted to these satisfied constraints and thus penalizes content that serves none of them. $F1_y$ is a balance between response precision and constraint recall. Further algorithm details about the counterexample generation and missing constraint update are referred to Appendix \ref{templates}.

\section{Experimental Evaluation}
\label{sec:experiments}

We evaluate ConQual-Audit by addressing three core research questions:

\textbf{RQ1 (Performance \& Cost)(\S\ref{sec:main-results}):} Does ConQual-Audit minimize evaluation false rates more effectively than reference-based, prompt-based LLM judges, and constraint evaluation baselines, and what is its computational trade-off? 

\textbf{RQ2 (The Generalizability of Constraint Audit)(\S\ref{sec:audit-effect}):} Can ConQual's qualification process be generalized as a model-agnostic filter to audit the constraints of other evaluators, and does it reduce their downstream decision errors? 

\textbf{RQ3 (Case Study)(\S\ref{sec:case-study}):} How does ConQual-Audit reshape decision boundaries to reflect both user preferences and task-grounded references? 

\begin{table*}[h]
\centering
\caption{Results averaged over LaMP-1, LaMP-2 and LaMP-4. Best in \textbf{bold}, second \underline{underlined}.}
\label{tab:main-results}
\resizebox{0.7\linewidth}{!}{%
\begin{tabular}{l cccccc rr}
\toprule
\textbf{Method} & \textbf{Pos.}$\uparrow$ & \textbf{Neg.}$\downarrow$ & \textbf{AUC}$\uparrow$ &
\textbf{FRR}$\downarrow$ & \textbf{FAR}$\downarrow$ & \makecell{\textbf{False Rate}}$\downarrow$ &
\makecell{\textbf{Tokens}}$\downarrow$ & \textbf{\#Cons.} \\
\midrule
BLEU-4      & 0.35 & 0.16 & 0.79 & 0.65 & 0.14 & 0.39 & 0.0 & -- \\
ROUGE-L     & 0.65 & 0.46 & 0.70 & 0.22 & 0.46 & 0.34 & 0.0 & -- \\
\midrule
BERTScore   & 0.95 & 0.92 & 0.74 & 0.00 & 1.00 & 0.50 & 28.3 & -- \\
MoverScore  & 0.65 & 0.57 & 0.76 & 0.08 & 0.79 & 0.44 & 29.0 & -- \\
\midrule
GEMBA       & 0.52 & 0.30 & 0.67 & 0.47 & 0.26 & 0.36 & 198.3   & -- \\
PerSE       & 0.41 & 0.26 & 0.66 & 0.62 & 0.20 & 0.41 & 1{,}212.6 & -- \\
FActScore   & 0.48 & 0.53 & 0.48 & 0.52 & 0.53 & 0.53 & 1{,}506.6 & -- \\
\midrule
ExPerT      & 0.07 & 0.00 & 0.54 & 0.92 & 0.01 & 0.46 & 3{,}027.0 & 3.47 \\
PREF        & 0.61 & 0.31 & \underline{0.84} & 0.26 & 0.25 & \underline{0.25} & 649.1 & 5.43 \\
\midrule
ConQual-Audit & 0.77 & 0.20 & \textbf{0.90} & 0.13 & 0.29 & \textbf{0.21} & 1{,}827.4 & 2.97 \\
\bottomrule
\end{tabular}
}
\end{table*}

\subsection{Experimental Setup}
\label{sec:exp-setup}

\textbf{Benchmarks.}
We evaluate on two benchmarks: (1) \emph{LaMP}~\cite{LaMP} (tasks 1, 2, and 4; 3{,}000 responses per task), testing semantic discrimination against surface form by evaluating 10 paraphrased positive answers, 10 negated answers, and 10 incorrect answers per user across 100 users; and (2) \emph{PrefEval}~\cite{zhao2025llmsrecognizepreferencesevaluating} (200 responses across 100 users), testing explicit preference compliance via paired Qwen3-14B~\cite{qwen3technicalreport} generations conditioned with and without the user's preference.

\textbf{Baselines.}
We compare against nine baselines across three families: (1) \emph{reference-based} metrics on LaMP (BLEU-4~\cite{papineni-etal-2002-bleu}, ROUGE-L~\cite{lin-2004-rouge}, BERTScore~\cite{bertscore}, MoverScore~\cite{MoverScore}); (2) \emph{LLM judges} (GEMBA~\cite{kocmi-federmann-2023-large}, PerSE~\cite{wang-etal-2024-learning-personalized}, FActScore~\cite{factscore}); and (3) \emph{constraint evaluators} (ExPerT~\cite{salemi2025expert}, PREF~\cite{fu2025prefreferencefreeevaluationpersonalised}). We also evaluate audited variants (ExPerT-Audit, PREF-Audit) via ConQual's qualification stage (\S\ref{subsec:qualification}) to test constraint audit generalizability.

\textbf{Metrics and Implementation.}
Responses are accepted at threshold $\tau \ge 0.5$. We report False Rejection Rate ($\text{FRR}$), False Acceptance Rate ($\text{FAR}$), their mean False Rate ($\text{FR}$), AUC, mean score margins, number of evaluated constraints, and end-to-end token costs. To isolate methodological gains from model capability, all LLM judges and ConQual modules ($G$, $V$, $J$) use Qwen3-8B~\cite{qwen3technicalreport}. ConQual-Audit uses $K = 2$ counterexample attempts, one gap-repair round per constraints, and at most three new constraints per repair.

\subsection{ConQual-Audit: Minimizing False Rate}
\label{sec:main-results}
\label{sec:prefeval-results}
\label{sec:lamp-results}

\emph{ConQual-Audit achieves the lowest overall false rate across both benchmarks.}
As shown in Tables~\ref{tab:main-results} and~\ref{tab:prefeval-results}, ConQual-Audit reduces overall classification error to $\text{FR} = 0.21$ on both datasets—a $16\%$ reduction over the strongest baseline on LaMP (PREF: $0.25$) and a $28\%$ reduction on PrefEval (PerSE: $0.29$)—while dominating in AUC ($0.90$ and $0.88$). This improvement stems from calibrated error control. On LaMP, ConQual-Audit halves the false rejection rate of PREF ($\text{FRR} = 13\%$ vs.\ $26\%$) at a comparable false acceptance rate ($\text{FAR} = 29\%$ vs.\ $25\%$), while evaluating roughly half as many constraints ($2.97$ vs.\ $5.43$). On PrefEval, it reduces false acceptances($\text{FAR} = 28\%$ vs.\ $45\%$ for PerSE) without inflating false rejections ($\text{FRR} = 15\%$ vs.\ $13\%$). Crucially, ConQual-Audit widens the score margin between positive and negative responses to $0.57$ on both benchmarks—nearly double any competing baseline ($\le 0.30$ on LaMP, $\le 0.32$ on PrefEval). Because audited constraints isolate necessary and sufficient conditions, valid paraphrases systematically satisfy them while invalid responses directly violate them.

\emph{Reference-based metrics confuses surface similarity with compliance.}
The acceptance sets $\widehat{\mathcal{A}}_{\mathrm{ref}}$ (Eq.~\eqref{eq:reference_induced_set}) exhibit severe failures: lexical metrics penalize benign stylistic rephrasing, falsely rejecting $65\%$ of valid answers on LaMP (BLEU-4 $\text{FRR} = 0.65$), whereas semantic embedding metrics mistake embedding proximity for alignment, accepting $100\%$ of negated or incorrect responses (BERTScore $\text{FAR} = 1.00$, MoverScore $0.79$).

\emph{Prompt-based LLM judges misplace the decision boundary.}
Unconstrained LLM judges relying on implicit boundaries $\widehat{\mathcal{A}}_{\mathrm{llm}}$ (Eq.~\eqref{eq:judge_induced_set}) fail to calibrate across task distributions. While GEMBA and PerSE show reasonable ranking capability on PrefEval (AUC $0.81$--$0.82$), they set overly permissive boundaries that accept $84\%$ and $45\%$ of non-compliant outputs ($\text{FAR}$). On LaMP, this behavior inverts into extreme conservatism, falsely rejecting $47\%$ and $62\%$ of correct answers ($\text{FRR}$). Decomposing responses into unweighted atomic facts (FActScore) fails entirely, hovering near chance on both benchmarks (AUC $0.48$--$0.52$).

\begin{table*}[h]
\centering
\caption{Results on PrefEval. Best in \textbf{bold}, second \underline{underlined}.}
\label{tab:prefeval-results}
\resizebox{0.7\linewidth}{!}{%
\begin{tabular}{l cccccc rr}
\toprule
\textbf{Method} & \textbf{Pos.}$\uparrow$ & \textbf{Neg.}$\downarrow$ & \textbf{AUC}$\uparrow$ &
\textbf{FRR}$\downarrow$ & \textbf{FAR}$\downarrow$ & \makecell{\textbf{False Rate}}$\downarrow$ &
\makecell{\textbf{Tokens}}$\downarrow$ & \textbf{\#Cons.} \\
\midrule
GEMBA            & 0.90 & 0.68 & \underline{0.82} & 0.03 & 0.84 & 0.44 & 670.1 & -- \\
PerSE            & 0.87 & 0.55 & 0.81 & 0.13 & 0.45 & \underline{0.29} & 1{,}889.2 & -- \\
FActScore        & 0.60 & 0.58 & 0.52 & 0.35 & 0.62 & 0.49 & 37{,}430.2 & -- \\
\midrule
ExPerT           & 0.08 & 0.06 & 0.53 & 0.97 & 0.04 & 0.50 & 9{,}663.9  & 8.23 \\
PREF             & 0.92 & 0.81 & 0.79 & 0.02 & 1.00 & 0.51 & 1{,}641.9  & 5.11 \\
\midrule
ExPerT-Audit     & 0.06 & 0.09 & 0.45 & 0.94 & 0.04 & 0.49 & 24{,}790.6 & 6.67 \\
PREF-Audit       & 0.91 & 0.74 & 0.78 & 0.03 & 0.89 & 0.46 & 16{,}006.7 & 3.64 \\
ConQual-Audit    & 0.80 & 0.23 & \textbf{0.88} & 0.15 & 0.28 & \textbf{0.21} & 12{,}628.5 & 2.36 \\
\bottomrule
\end{tabular}
}
\end{table*}

\emph{Constraint criteria violate necessity and sufficiency.}
Constraint baselines introduce high false rates by generating criteria that either over- or under-specify the solution space. ExPerT conditions constraints strictly on reference tokens; because gold references are often abstract criteria (PrefEval) or brief sentences (LaMP), valid candidate answers naturally omit arbitrary reference details. Its constraint sets thus violate necessity (Eq.~\eqref{eq:necessity}), falsely rejecting $92\%$ (LaMP) and $97\%$ (PrefEval) of valid outputs ($\text{FRR}$). Conversely, PREF constructs generic rubric criteria easily satisfied by standard generation quality regardless of preference adherence (mean negative score $0.81$); its constraints violate sufficiency (Eq.~\eqref{eq:sufficiency}), accepting every negative response on PrefEval ($\text{FAR} = 1.00$). Both evaluate more than twice as many constraints as ConQual-Audit ($5.11$--$8.23$ vs.\ $2.36$), yet yield the highest false rates ($\text{FR} \ge 0.50$).

\emph{Inference cost and amortized verification.}
On PrefEval, ConQual-Audit consumes 12.6k tokens per response end-to-end, higher than single-prompt judges (GEMBA: 0.7k, PerSE: 1.9k), but one-third of atomic-decomposition methods (FActScore: 37.4k) and substantially less than auditing unvetted external criteria (ExPerT-Audit: 24.8k, PREF-Audit: 16.0k). Over $80\%$ of this compute ($\sim 10.1\text{k}$ tokens) is front-loaded into one-time constraint synthesis and qualification per instruction. The runtime verification step requires only 2.5k tokens per candidate response, readily amortizing across multiple model outputs ($N > 1$). On LaMP, end-to-end cost scales down to 1.8k tokens per response, falling between PREF (0.6k) and ExPerT (3.0k). This focused compute investment directly delivers the lowest ($\text{FR} = 0.21$ across both benchmarks.

\subsection{Generalization of Constraint Audit}
\label{sec:audit-effect}
Because ConQual's audit (\S\ref{subsec:qualification}) operates over the explicit constraints, it can function as a model-agnostic filter for other constraint-based evaluators. To test its portability, we apply the audit to the response aspects generated by ExPerT and the rubric criteria produced by PREF, re-scoring responses with each method's native verifier under both original and audited constraints (Table~\ref{tab:prefeval-results}). Table~\ref{tab:audit-impact} details the resulting constraint reductions, classification AUC, and false rates.

\begin{table}[h]
\centering
\caption{Auditing the constraints of ExPerT and PREF.}
\label{tab:audit-impact}
\resizebox{0.85\linewidth}{!}{%
\begin{tabular}{ccccc}
\toprule
 & \multicolumn{2}{c}{\textbf{ExPerT}} & \multicolumn{2}{c}{\textbf{PREF}} \\
\cmidrule(lr){2-3} \cmidrule(lr){4-5}
 & Orig. & Audit & Orig. & Audit \\
\midrule
\#Cons.          & 8.23 & 6.67 & 5.11 & 3.64 \\
Removed / added  & \multicolumn{2}{c}{1.78 / 0.22} & \multicolumn{2}{c}{1.51 / 0.05} \\
AUC $\uparrow$   & 0.53 & 0.45 & 0.82 & 0.78 \\
FRR $\downarrow$ & 0.96 & 0.94 & 0.02 & 0.03 \\
FAR $\downarrow$ & 0.04 & 0.04 & 1.00 & 0.89 \\
FR $\downarrow$  & 0.50 & 0.49 & 0.51 & 0.46 \\
\midrule
Reproduced       & \multicolumn{2}{c}{99.8\%} & \multicolumn{2}{c}{77\%} \\
\bottomrule
\end{tabular}
}
\end{table}

\emph{Eliminating false constraints can reduce the false rate.}
The audit systematically identifies and removes unnecessary or over-constraint constraints: it prunes 1.78 of ExPerT's 8.23 aspects ($-19\%$) and 1.51 of PREF's 5.11 criteria ($-29\%$), while rarely adding new ones ($+0.22$ and $+0.05$, respectively). Operating on these audited constraint sets reduces overall classification error: the false rate (FR) drops from 0.50 to 0.49 for ExPerT and from 0.51 to 0.46 for PREF. For PREF, this improvement directly addresses its primary failure mode identified in Table~\ref{tab:prefeval-results}, reducing its false acceptance rate (FAR) down from 1.00 to 0.89 while keeping false rejections at a very low level (FRR: 0.02 $\to$ 0.03).

\begin{figure*}[h]
    \centering
    \includegraphics[width=1\linewidth]{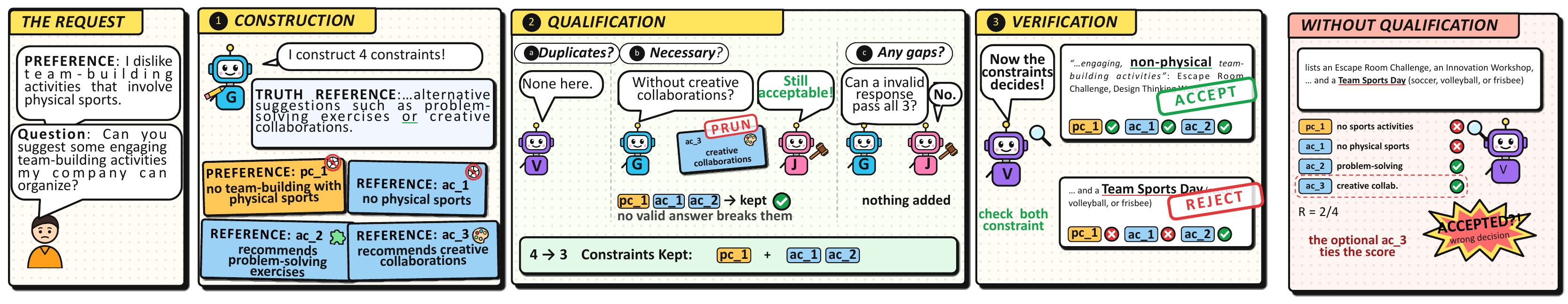}
    \caption{ConQual-Audit. (1) The generator $G$ derives one preference constraint and three answer constraints from the user preference and the truth-grounded reference (if verify directly, they accept the negative response ($R = 2/4 = \tau$)). (2) Qualification prunes \texttt{ac\_3}, which the reference makes optional, and finds no duplicate and no gap. (3) The verifier $V$ checks responses against the qualified constraint set: the positive response is accepted and the negative one rejected.}
    \label{fig:case-pipeline}
\end{figure*}

\emph{Decouple constraint quality from evaluation randomness.}
To isolate the intrinsic value of constraint pruning from downstream evaluator noise, we analyze a counterfactual setting where verifier decisions on unchanged constraints remain fixed. Under this controlled ablation, removing audited constraints strictly improves classification performance: PREF's AUC increases from 0.82 to 0.84 (FR drops to 0.45), while ExPerT's FR falls from 0.50 to 0.47 at a steady 0.53 AUC. In the end-to-end re-evaluation reported in Table~\ref{tab:audit-impact}, these gains are masked by test-retest instability in the underlying LLM judges: PREF's verifier reproduces only 77\% of its decisions on identical criteria across runs, introducing a baseline drift of $\pm 0.03$ AUC that accounts for the nominal drop to 0.78. Thus, while the audit effectively cleans the constraint space, end-to-end performance remains constrained by the variance of third-party scoring backends.

\emph{Inference overhead and ConQual's efficiency advantage.}
The primary trade-off of the audit process is computational cost. Because counterexample generation and verification consume 5--6k tokens per constraint, overhead scales linearly with the size of the unvetted constraint set. Auditing ExPerT (8.7 aspects per response) and PREF (5.1 criteria per prompt) inflates end-to-end inference to 24.8k and 16.0k tokens per response, respectively—overhead that is only marginally offset by downstream scoring speedups (8\%--25\% fewer verification tokens). By contrast, ConQual constructs and audits a compact set of only 3.9 global constraints once per question. Consequently, ConQual-Audit achieves higher accuracy (AUC 0.88, FR 0.21) at lower total compute (12.6k tokens). The audit thus functions as an effective off-the-shelf refiner for third-party criteria, but delivers its best cost-accuracy profile when integrated into ConQual's unified pipeline.

\subsection{ConQual-Audit: Case Study}
\label{sec:audit-case}
\label{sec:case-study}

Figure~\ref{fig:case-pipeline} illustrates a case study from PrefEval benchmark of ConQual-Audit. How it reshapes the decision boundary to jointly enforce user-specified preferences and task-grounded references.

The user specifies a negative preference against 'physical sports 'in team-building activities, while the gold reference suggests 'non-physical alternatives', such as 'problem-solving exercises' or 'creative collaborations'. The construction produces four candidate constraints: a user-specified preference against physical sports (\texttt{pc\_1}) with three task-grounded reference criteria (\\texttt{ac\_1}--\texttt{ac\_3}).  

Without constraint qualification, an invalid candidate proposal for a 'Team Sports Day' satisfied two constraints (\texttt{ac\_2}, \texttt{ac\_3}) but violated two other sports-related constraints (\texttt{pc\_1}, \texttt{ac\_1}), and could be thus falsely accepted. This problem is resolved by ConQual-Audit qualification with counterexamples: because omitting creative collaborations (\texttt{ac\_3}) leaves a response fully acceptable, \texttt{ac\_3} is pruned as unnecessary. Evaluated on the qualified three-constraint set (\texttt{pc\_1}, \texttt{ac\_1}, \texttt{ac\_2}), the invalid candidate is identified and rejected, while valid candidates are correctly accepted.

\section{Conclusion}
In this work, we presented {ConQual-Audit}, an audit-based framework for evaluating LLM personalization through semantic constraint construction, qualification, and verification. {ConQual-Audit} synthesizes candidate constraints grounded jointly in user preferences and task references, systematically auditing them for individual necessity and joint sufficiency, pruning unnecessary constraints to minimize false rejections while repairing missing constraints to prevent false acceptances. By evaluating candidate responses via the harmonic mean of constraint satisfaction and response attribution, {ConQual-Audit} establishes a calibrated decision boundary that resolves both over- and under-constrained failure modes.

Empirical evaluations across PrefEval and LaMP show that {ConQual-Audit} achieves state-of-the-art performance, attaining top AUC scores ($0.88$ and $0.90$) and the lowest overall false rate ($\text{FR} = 0.21$ on both datasets) while operating on the most compact constraint sets ($2.36$ and $2.97$ per query). Although constraint qualification requires an upfront token investment, this cost readily amortizes during runtime verification across multiple candidate generations. Furthermore, our case study demonstrates that qualification structurally realigns the decision boundary: pruning unqualified constraint details prevents auxiliary criteria from diluting core constraints and triggering false acceptances.

Finally, our findings confirm that constraint qualification is model-agnostic and portable: applying the audit to existing evaluators (ExPerT and PREF) systematically refines their constraint sets and consistently lowers downstream false rates. These results demonstrate that unvetted constraint extraction routinely dilutes explicit user preferences with superficial task artifacts, establishing systematic constraint auditing as a principled foundation for reliable personalization evaluation.

\section*{Acknowledgments}
AI tools were used solely for language refinement. All scientific contributions are authors' original work. This research is supported by the Technology Innovation Institute, Abu Dhabi, UAE.

\section*{Limitations}
ConQual-Audit depends on the models that play the generator, the verifier and the judge. In our experiments a single Qwen3 model fills all three roles, so their errors are correlated. Qualification is also costly. It raises the cost per PrefEval response from 3.4k to 12.6k tokens end to end, and ConQual-Audit has the highest online time of the evaluators we compare (27\,s per response). The framework needs a reference answer and an explicit preference to construct constraints, so it does not apply to open-ended requests that lack them. It scores a response by the share of satisfied constraints, all weighted equally, at a fixed threshold. Finally, our evaluation covers English benchmarks with constructed positive and negative responses. Results may differ for other domains, languages and response distributions.

\bibliography{main}

\clearpage
\appendix
\section{LLM Instruction Templates}
\label{templates}
\label{Templates}

\begin{algorithm}[h]
\caption{Functions of Algorithm~\ref{alg:qualify}. \textsc{Adverse} returns a response that satisfies every constraint
in $A$ and violates every constraint in $B$ (or, with $B = \emptyset$, breaks requirement $r$), or $\bot$ after $K$
attempts. \textsc{Update} asks $G$, given $J$'s explanation $J(\tilde{y})$, for at most three new constraints of the kind
that $r$ calls for and returns those that exclude $\tilde{y}$ (and, for answer constraints, hold for $y^{\ast}$).}
\label{alg:qualify-functions}
\DontPrintSemicolon
\LinesNumbered
\SetFuncSty{textsc}
\SetArgSty{textup}
\SetKwProg{Fn}{Function}{:}{}
\SetKwFunction{FW}{Adverse}
\SetKwFunction{ADD}{Update}
\KwIn{$q, H, y^\ast$ and $G, J, V$ as in Algorithm~\ref{alg:conqual_p}; attempts $K$}
\BlankLine
\Fn{\FW{$A, B, r$}}{
  \For{$k = 1, \dots, K$}{
    $\tilde{y} \gets G(q, A, B, r)$
    \lIf{$\forall c \in A: V(\tilde{y}, c) = \textsc{E}$ \textbf{and} $\forall c \in B: V(\tilde{y}, c) \in \{\textsc{C}, \textsc{N}\}$}{\Return{$\tilde{y}$}}
  }
  \Return{$\bot$}\;
}
\BlankLine
\Fn{\ADD{$\mathcal{C}, \tilde{y}, r$}}{
  $M \gets y^\ast$ \textbf{if} $r = y^\ast$ \textbf{else} $H$\tcp*{reference material of $r$}
  $\mathcal{C}^{+} \gets G(q, \mathcal{C}, \tilde{y}, J(\tilde{y}), M)$\tcp*{$|\mathcal{C}^{+}| \leq 3$, kind of $r$}
  $\mathcal{C}^{+} \gets \{c \in \mathcal{C}^{+} : V(\tilde{y}, c) \in \{\textsc{C}, \textsc{N}\}\}$\;
  \lIf{$r = y^\ast$}{$\mathcal{C}^{+} \gets \{c \in \mathcal{C}^{+} : V(y^\ast, c) = \textsc{E}\}$}
  \Return{$\mathcal{C}^{+}$}\;
}
\end{algorithm}

\paragraph{Adversarial generation.}
Both audits rest on counterexamples written by the generator $G$ (\textsc{Adverse} in Algorithm~\ref{alg:qualify-functions}). $G$ receives the query $q$, a set $A$ of constraints the response must satisfy, and a target to break: either a set $B$ of constraints (over-constraint pruning, $B = \{c\}$) or an alignment requirement $r \in \{p, y^{\ast}\}$ (gap repair, $B = \emptyset$, where the response must break the stated preference or disagree with the reference). It writes a complete, plausible answer of at most 200 words. The verifier $V$ then checks the response against $A \cup B$, and it is used only if $V$ labels every constraint in $A$ as entailed and every constraint in $B$ as contradicted or neutral (a neutral label means the response omits the required content, so it does not satisfy the constraint). Otherwise $G$ tries again, up to $K = 2$ times; if no attempt passes, \textsc{Adverse} returns $\bot$ and the constraint set is left unchanged. A valid counterexample is then judged by $J$, which decides separately whether it is aligned with the preference and with the reference and explains each decision.

\paragraph{Constraint update.}
When $J$ finds that a gap-repair counterexample $\tilde{y}$ satisfies every current constraint yet breaks a requirement $r$, the constraint set admits a misaligned response, and \textsc{Update} closes the gap for each broken requirement. $G$ receives $q$, the current constraints $\mathcal{C}$, $\tilde{y}$, $J$'s explanation of the failure, and the material $r$ rests on: the reference $y^{\ast}$ when $\tilde{y}$ disagrees with it, or the user's stated preferences $H$ when it breaks them. It proposes at most three new constraints of the matching kind (answer constraints from $y^{\ast}$, preference constraints from $H$) that $\tilde{y}$ violates, that every acceptable response would satisfy, and that do not restate a current constraint. A proposed constraint is added only if $V$ confirms that it excludes $\tilde{y}$ (contradicted or neutral); an answer constraint must also be entailed by $y^{\ast}$ itself, so that the repair never rejects the reference. Each requirement gets one repair round.

\end{document}